%% file: main.tex
\pdfoutput=1

\documentclass[final]{cvpr}

\usepackage{times}
\usepackage{epsfig}
\usepackage{graphicx}
\usepackage{amsmath}
\usepackage{amssymb}

\usepackage{verbatim}
\usepackage{color,soul}
\usepackage{dblfloatfix}
\usepackage{stmaryrd}
\usepackage{caption} 
\usepackage[pagebackref=true,breaklinks=true,colorlinks,bookmarks=false]{hyperref}

\def\cvprPaperID{4318} 
\def\confYear{CVPR 2021}

\newif\ifhidepagenumbers

\iftoggle{cvprfinal}{
    \ifhidepagenumbers
        \pagestyle{empty}
    \fi
}

\begin{document}

\newif\ifisolatesupplementary 
\ifisolatesupplementary 
\newcommand{\suppref}[2]{{\color{red}{#2}}}
\setcounter{table}{1}
\setcounter{figure}{6}
\else
\newcommand{\suppref}[2]{\ref{#1}}
\fi

\ifisolatesupplementary
\else
\title{BRepNet: A topological message passing system for solid models}

\author{
    Joseph G. Lambourne\\
    Autodesk Research \\
    \and
    Karl D.D. Willis\\
    Autodesk Research
    \and
    Pradeep Kumar Jayaraman\\
    Autodesk Research
    \and
    Aditya Sanghi\\
    Autodesk Research
    \and
    Peter Meltzer\\
    UCL, Computer Science
    \and
    Hooman Shayani\\
    Autodesk Research
}

\maketitle
\iftoggle{cvprfinal}{
    \ifhidepagenumbers
        \thispagestyle{empty}
    \fi
}

\begin{abstract}
Boundary representation (B-rep) models are the standard way 3D shapes are described  in Computer-Aided Design (CAD) applications.  They combine lightweight parametric curves and surfaces with topological information which connects the geometric entities to describe manifolds.  In this paper we introduce BRepNet, a neural network architecture designed to operate directly on B-rep data structures, avoiding the need to approximate the model as meshes or point clouds.   
BRepNet defines convolutional kernels with respect to oriented coedges in the data structure. In the neighborhood of each coedge, a small collection of faces, edges and coedges can be identified and patterns in the feature vectors from these entities detected by specific learnable parameters.
In addition, to encourage further deep learning research with B-reps, we publish the \textit{Fusion 360 Gallery} segmentation dataset.  A collection of over 35,000 B-rep models annotated with information about the modeling operations which created each face.  We demonstrate that BRepNet can segment these models with  higher accuracy than methods working on meshes, and point clouds. 
\end{abstract}
\fi

\newif\ifshowcomments
\showcommentstrue
\ifshowcomments
\newcommand{\joe}[1]{{\color{red}{[Joe: #1]}}}
\newcommand{\karl}[1]{{\color{purple}{[Karl: #1]}}}
\newcommand{\pradeep}[1]{{\color{cyan}{[Pradeep: #1]}}}
\newcommand{\aditya}[1]{{\color{green}{[Aditya: #1]}}}
\newcommand{\pete}[1]{{\color{yellow}{[Pete: #1]}}}
\newcommand{\hooman}[1]{{\color{blue}{[Hooman: #1]}}}
\else
\newcommand{\joe}[1]{}
\newcommand{\karl}[1]{}
\newcommand{\pradeep}[1]{}
\newcommand{\aditya}[1]{}
\newcommand{\pete}[1]{}
\newcommand{\hooman}[1]{}
\fi

\newcommand{\R}{\mathbb{R}}
\newcommand{\boldsubsec}[1]{~\\\noindent\textbf{#1}}

\ifisolatesupplementary 
\else
\input{sec/introduction}
\input{sec/related_work}
\input{sec/method}

\input{sec/data}
\input{sec/experiments}
\input{sec/conclusions.tex}
\fi

\ifisolatesupplementary 
\input{sec/appendix.tex}

\fi


\bibliography{main}

\ifisolatesupplementary 
\else
\input{sec/appendix.tex}
\fi

\end{document}

%% file: sec/introduction.tex
\begin{figure}
    \centering
    \includegraphics[width=\columnwidth]{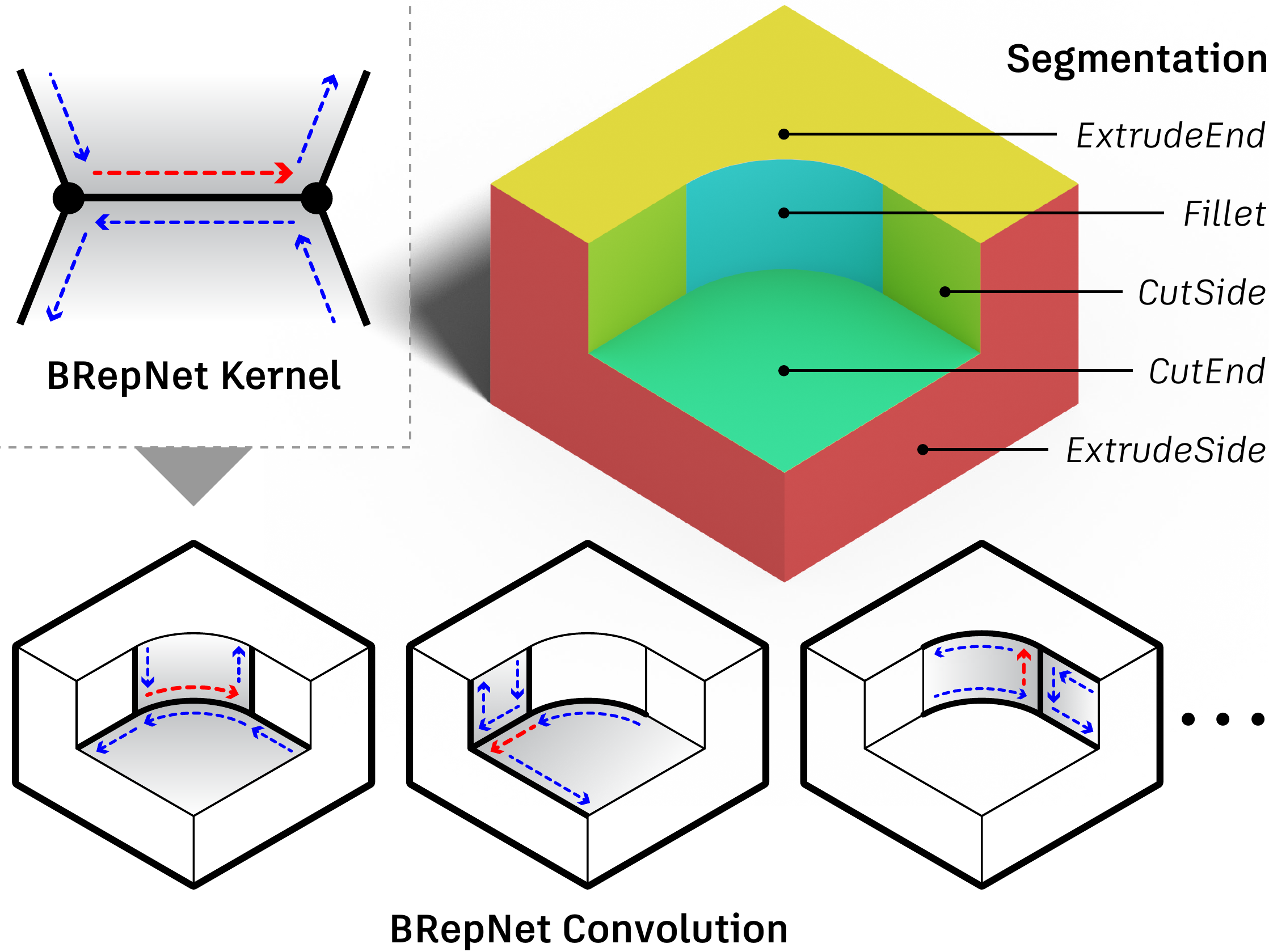}
    \caption{
    BRepNet convolutional kernels are defined with respect to topological entities called \emph{coedges} (dashed arrows).  Feature vectors from a small collection of faces (grey), edges (black) and coedges (blue) adjacent to each coedge (red) are multiplied by the learnable parameters in the kernel. The hidden states arising from the convolution can then be pooled to perform face segmentation.}
    \label{figure:teaser}
\end{figure}

\section{Introduction}

Boundary representation (B-rep) models are the de facto standard for describing 3D objects in commercial Computer Aided Design (CAD) software. They consist of collections of trimmed parametric surfaces along with the adjacency relationships between them \cite{Weiler1986}. Prismatic shapes can be represented using lightweight primitive curves and surfaces while free-form objects can be defined using NURBS \cite{PiegTill96}.  Although this makes the representation both compact and expressive, the complexity of the data structures and limited availability of labelled datasets has presented a high barrier to entry for researchers. 

The problem of segmenting B-rep models, based on learned patterns, is of particular interest as it allows the automation of many laborious manual tasks in CAD, Computer Aided Engineering (CAE) and Computer Aided Process Planning (CAPP) \cite{Babic2008, Yue2002, Alwswasi2018, Shi2020}. Currently these require a user to repeatedly select groups of faces and/or edges as input for the modeling or manufacturing operation. Examples include model simplification in preparation for finite element analysis \cite{Danglade2014} and segmenting a model according to the manufacturing process or machining toolpath strategy required to make the object \cite{Alwswasi2018, Yue2002}.  

In addition, parametric feature history is often lost when models are exchanged between different CAD applications \cite{ByungChul2007} and many commercial CAD systems use segmentation algorithms to recover this information  \cite{inventorFR, featureWorks}.

Although attempts were made in the 90s to apply neural networks to the task of B-rep segmentation \cite{Henderson1994, Chen1998, Marquez1999, Ding2004, Sunil2009, Shi2020}, the absence of machine learning frameworks and large labelled datasets caused progress to stall until very recently \cite{jayaraman2020uvnet,cao2020}. 
In this paper we introduce BRepNet, a novel neural network architecture designed specifically to operate directly on the faces and edges of B-rep data structures and take full advantage of the topological relationships between them. In addition, we hope to revitalize interest in the problem of B-rep segmentation with the publication of the \textit{Fusion 360 Gallery} segmentation dataset. For the first time we provide a collection of over 35,000 3D models, in multiple representations, annotated with segmentation labels revealing the modeling operations used to create them.   

The BRepNet approach is motivated by the observation that in convolutional neural networks for image processing, the weights operate on pixels with known locations within the filter window.  A similar arrangement can be achieved with B-reps, where a small collection of faces, edges and coedges can be identified at well defined locations relative to each coedge in the data structure (see Figure \ref{figure:teaser}). Feature vectors can be extracted from these neighbouring entities and concatenated in a known order, allowing convolution to take place as a matrix/vector multiplication \cite{Hanocka2019,Jia2014LearningSI}. As in image convolution, specific entities relative to each coedge map to specific learnable parameters in our convolutional kernels, allowing patterns in the input data to be easily recognized \cite{Niepert2016,Bouritsas_2019}. The key contributions are as follows:
\begin{itemize}
    \item We introduce BRepNet, a network architecture using a novel convolution technique which takes full advantage of the topological information the B-rep stores. 
    \item We publish the \textit{Fusion 360 Gallery} segmentation dataset that contains over 35,000 segmented 3D models in B-rep, mesh, and point cloud format.
    \item We provide experimental results on the \textit{Fusion 360 Gallery} segmentation task, including ablation studies and comparisons to other representations and methods.
\end{itemize}
Our results demonstrate that direct use of B-rep data  solves the \textit{Fusion 360 Gallery} segmentation problem with higher performance and parameter efficiency than other techniques based on point cloud and mesh representations.

%% file: sec/related_work.tex
\section{Related work}
Historically, the task of B-rep segmentation has focused on the detection of form features (connected regions of a model with a characteristic shape or pattern with some significance \cite{Shi2020}).  Feature detection has been an active area of research since the mid 1970s \cite{Henderson1994}, with a range of different heuristic approaches investigated \cite{Kyprianou1980, Staley1983, Ansaldi1985, Joshi1988GraphbasedHF, Sakurai1990, Ma2010AutomaticDO}. 
\\

\noindent\textbf{Early neural networks.} 
Neural networks were first employed by Prabhakar \etal \cite{Prabhakar1992} with a number of extensions and refinements made over the years \cite{nezis1997, Ding2004, Sunil2009}.  In these early works the B-rep structure is first converted to a face adjacency graph with node features extracted from the B-rep faces and attributes for the arcs extracted from the B-rep edges.  Heuristics are then used to break the graph into small connected components which are passed to the networks individually.  These techniques were limited by the computer power of the time and so the networks see only a small part of the B-rep at once.

\boldsubsec{Voxels.} Feature detection methods based on voxels \cite{Balu2016, Zhang2018} offer some advantages for manufacturability analysis, however the cubic storage complexity puts severe limitations on the size of geometric features which can be detected. As CAD models often contain small but important features, the applicability of these techniques with current GPU hardware is quite limited.

\boldsubsec{Point clouds.}  Point cloud segmentation has shown excellent results in recent years \cite{qi2017pointnet, qi2017pointnet++, WangDGCNN2018}, but typically requires a large number of points to be uniformly sampled from the (B-rep) objects' surface.  Faces with small areas can easily be under-sampled and incorrectly classified.

\boldsubsec{Meshes.}   
Triangle meshes are another important representation for 3D objects, and a number of authors have proposed convolution strategies which operate on them  \cite{Wang2018,Hanocka2019,Bouritsas_2019, Liu2020}. 
MeshCNN \cite{Hanocka2019} operates on the edges of a triangle mesh with convolution carried out by aggregating information from the five edges of two adjacent triangles onto the central edge.
Liu \etal \cite{Liu2020} introduce a convolution scheme which operates on directed triangle edges and use it to generate neural network conditioned subdivision surfaces.  Although the data structures for triangle meshes are simple, converting B-reps to high quality manifold meshes requires special meshing procedures.  By working directly on the original B-rep topology we can avoid the requirement to generate good quality meshes and operate directly on a more compact representation.  

\boldsubsec{Graphs.} B-rep model segmentation can also be viewed as a node classification problem on graphs. Two concurrent unpublished works have applied graph convolution approaches to B-rep segmentation \cite{jayaraman2020uvnet, cao2020}.   Jayaraman \etal \cite{jayaraman2020uvnet} uses convolution layers to create input features from grids of 3D points and normal vectors, while Cao \etal \cite{cao2020} uses only planar faces which can be directly represented as feature vectors of length 4.  In both cases the B-rep data structure is translated to a face adjacency graph which causes some information about relative topological locations of nearby entities to lost.

%% file: sec/method.tex
\begin{figure*}[!t]
     \includegraphics[width=1\linewidth]{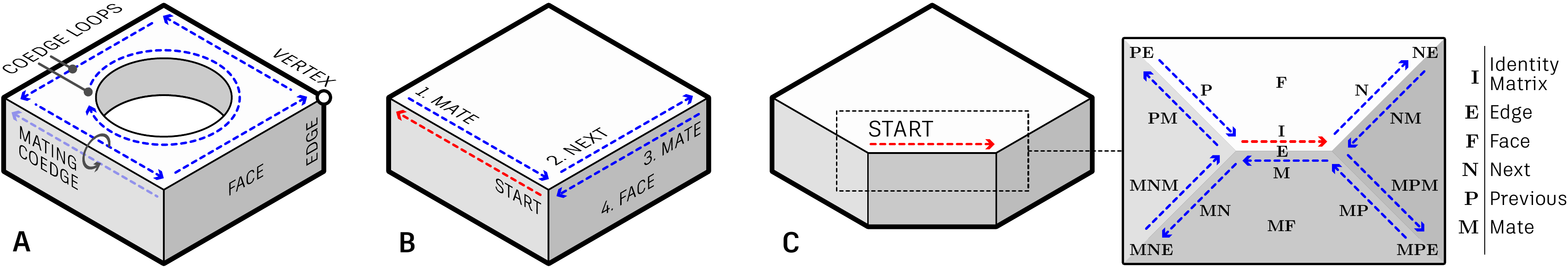}
     \caption{A) B-rep topology comprises faces, edges, loops, coedges and vertices. B) Starting from a given coedge (red), the topology can be traversed by following a sequence of instructions which indicate which entity to move to in the next hop.  The instruction sequence \emph{\{mate, next, mate, face\}} is illustrated. C) The walks from the red coedge to some neighbouring entities are described in terms of products of the incidence matrices $\mathbf{N}$, $\mathbf{P}$, $\mathbf{M}$, $\mathbf{F}$ and $\mathbf{E}$.}
     \label{fig:topological_walk}
\end{figure*}

\section{Method}
\label{section:method}

\subsection{B-rep data structures}
\label{section:solid_modeling_datas_tructures}
Industrial CAD packages have internal data structures which are similar to the partial entity structure described by Lee \etal \cite{Lee2001}.  These structures support the modeling of 2-dimensional manifolds, 3-dimensional volumes and even non-manifold complexes which can arise as intermediate states in boolean operations \cite{Weiler1986}.   

A B-rep comprises of faces, edges, loops, coedges and vertices (Figure~\ref{fig:topological_walk}a).  A face is a connected region of the model's surface which may have internal holes \cite{Eastman1979}.  An edge defines the curve where two faces meet and a vertex defines the point where edges meet. Faces have an underlying parametric surface which is divided into visible and hidden regions by a series of boundary loops.  

Each loop consists of a doubly linked list of directed edges called coedges,  topological entities which are used to represent the adjacency relationships in the B-rep \cite{Lee2001}.  A coedge stores pointers to the next and previous coedge in the loop, its adjacent or ``mating" coedge, its parent face and parent edge. In this work we consider only closed and manifold B-reps where each coedge has exactly one mating coedge, providing sufficient information for the edges in the structure to be traversed in the same way as in the winged edge \cite{Baumgart1972} and QuadEdge \cite{Guibas1983} data structures.

\subsection{Topological walks}

By following the pointers which the coedges store, we can walk from a given coedge on the B-rep to entities in its neighborhood.  The choice of which pointer to follow at each hop can be thought of as a sequence of instructions which will take us from some starting coedge to a destination coedge.  From there we can optionally make one final jump to its owner edge or face. This sequence of instructions defines a topological walk.

An example of a simple topological walk for the instruction sequence: \emph{\{mate, next, mate, face\}} is shown in Figure~\ref{fig:topological_walk}b. The starting coedge is shown in red and the coedges traversed during the walk are shown in blue. 

For a set of B-rep faces ${\mathbf{f}=\{f_1, f_2, ..., f_{|\mathbf{f}|}\}}$, edges ${\mathbf{e}=\{e_1, e_2, ..., e_{|\mathbf{e}|}\}}$,  and coedges ${\mathbf{c}=\{c_1, c_2, ..., c_{|\mathbf{c}|}\}}$, geometric information can be extracted and used to build three input feature matrices ${\mathbf{X^f} \in \R^{|\mathbf{f}| \times p}}$, ${\mathbf{X^e} \in \R^{|\mathbf{e}| \times q}}$ and ${\mathbf{X^c} \in \R^{|\mathbf{c}| \times r}}$ for the face features, edge features and coedge features respectively as described in Section \ref{section:input_features}. 

The next, previous and mating adjacency relationships between coedges can be written as three matrices:

\begin{equation}
\mathbf{N}, \mathbf{P}, \mathbf{M} \in \{0,1\}^{|\mathbf{c}| \times |\mathbf{c}|}
\end{equation}

Here $\mathbf{N}_{ij}=1$ indicates that $c_j$ is the next coedge in the loop from $c_i$ and $\mathbf{M}_{ij}=1$ when coedge $c_j$ is the mate of coedge $c_i$.  As each coedge has exactly one next, previous and mating coedge, these matrices simply define permutations on the list of coedges in the B-rep. Also we can see that $\mathbf{P}=\mathbf{N}^{-1}=\mathbf{N}^{T}$.  A matrix defining a topological walk between two coedges can then be built by multiplying $\mathbf{N}$, $\mathbf{P}$ and $\mathbf{M}$ in the sequence in which the $next$, $previous$ and $mate$ instructions appear in the walk (Figure~\ref{fig:topological_walk}c).  

The relationships between a coedge and its parent face and parent edge can also be represented using incidence matrices $\mathbf{F} \in \{0,1\}^{|\mathbf{c}| \times |\mathbf{f}|}$ and $\mathbf{E} \in \{0,1\}^{|\mathbf{c}| \times |\mathbf{e}|}$. Here  $\mathbf{F}_{ij}=1$ indicates that coedge $c_i$ is in a loop around face $f_j$ and $\mathbf{E}_{ij}=1$ indicates that coedge $c_i$ belongs to edge $e_j$.

The transform $\mathbf{\Psi} = \mathbf{F}\mathbf{X^f}$allow us to construct a matrix $\mathbf{\Psi} \in \R^{|\mathbf{c}| \times p}$ by copying the $i$th row of the matrix of face features $\mathbf{X^f}$ to the $j$th row of $\mathbf{\Psi}$ for each coedge $c_j$ with parent face $f_i$.
The matrix  $\mathbf{E}$ works in a similar way for edges.  Topological walks which terminate on faces or edges can then be represented in matrix form by multiplying the matrix for the walk over the coedges by $\mathbf{E}$ or $\mathbf{F}$.

\subsection{Input feature extraction}
\label{section:input_features}
Geometric feature information from the faces, edges and coedges of the B-rep are passed into the network in the feature matrices ${\mathbf{X^f}}$, ${\mathbf{X^e}}$ and ${\mathbf{X^c}}$.  One  approach to the extraction of geometric features from B-rep faces is given by \cite{jayaraman2020uvnet}, where grids of points and normal vectors are sampled from the parametric surface geometry and compressed into feature vectors using a CNN. In this work we investigate whether the \textit{Fusion 360 Gallery} segmentation problem can be solved without providing the network with any coordinate information, instead using only a small amount of extremely concise information from the B-rep data structure.  Using coordinate free input features has the advantage that they are invariant to translation and rotation and protects the intellectual property of CAD operators by not reveling the model geometry, while still allowing the network to perform useful tasks.

For face features, the network is given a one-hot vector encoding of the possible surface types (plane, cylinder, cone, sphere, torus).  One additional value is used to indicate a rational NURBS surface \cite{PiegTill96}.  In the case of non-rational B-splines all these values will be zero.   We also provide the network with the area of each face. For edge features we provide a one-hot vector encoding of the possible kinds of edge geometry (line, circle, ellipse, helix, intersection curve).  We encode edge convexity in three one-hot values (concave edge, convex edge, smooth edge).   One additional flag indicates if an edge forms a closed loop.  Finally the edge length is added.  
For coedges, the network is passed a single flag indicating whether the direction of the coedge is the same as the direction of the parametric curve of the edge.  The input features are standardized over the training set and the same scaling applied to the validation and test sets.  More detail is in the supplementary material. 

\subsection{Convolution}
\label{section:convolution}
Convolutional kernels in BRepNet are defined relative to the coedges of the B-rep.  As noted by Lui \etal \cite{Liu2020}, because coedges are directed this removes the ambiguity between the faces to the left and right of a coedge and avoids the need to aggregate features using symmetric functions as in \cite{Hanocka2019}.  The relative topological locations between a starting coedge and the faces, edges and coedges which will take part in a convolution are defined by topological walks.  Each walk can be expressed in matrix form by multiplying the matrices $\mathbf{N}$, $\mathbf{P}$, $\mathbf{M}$, $\mathbf{F}$ and $\mathbf{E}$ in the order in which the \emph{next, previous, mate, face} or \emph{edge} instructions must be executed.  An example of a collection of faces, edges and coedges which can be used in a BRepNet kernel is shown in Figure \ref{fig:topological_walk}c.  The products of the matrices required to reach each of the destination entities are marked.  The matrices encoding the walks to faces, edges and coedges are arranged in three lists  $K^f$, $K^e$ and $K^c$ receptively. 

\begin{figure}[t]
    \centering
    \includegraphics[width=\columnwidth]{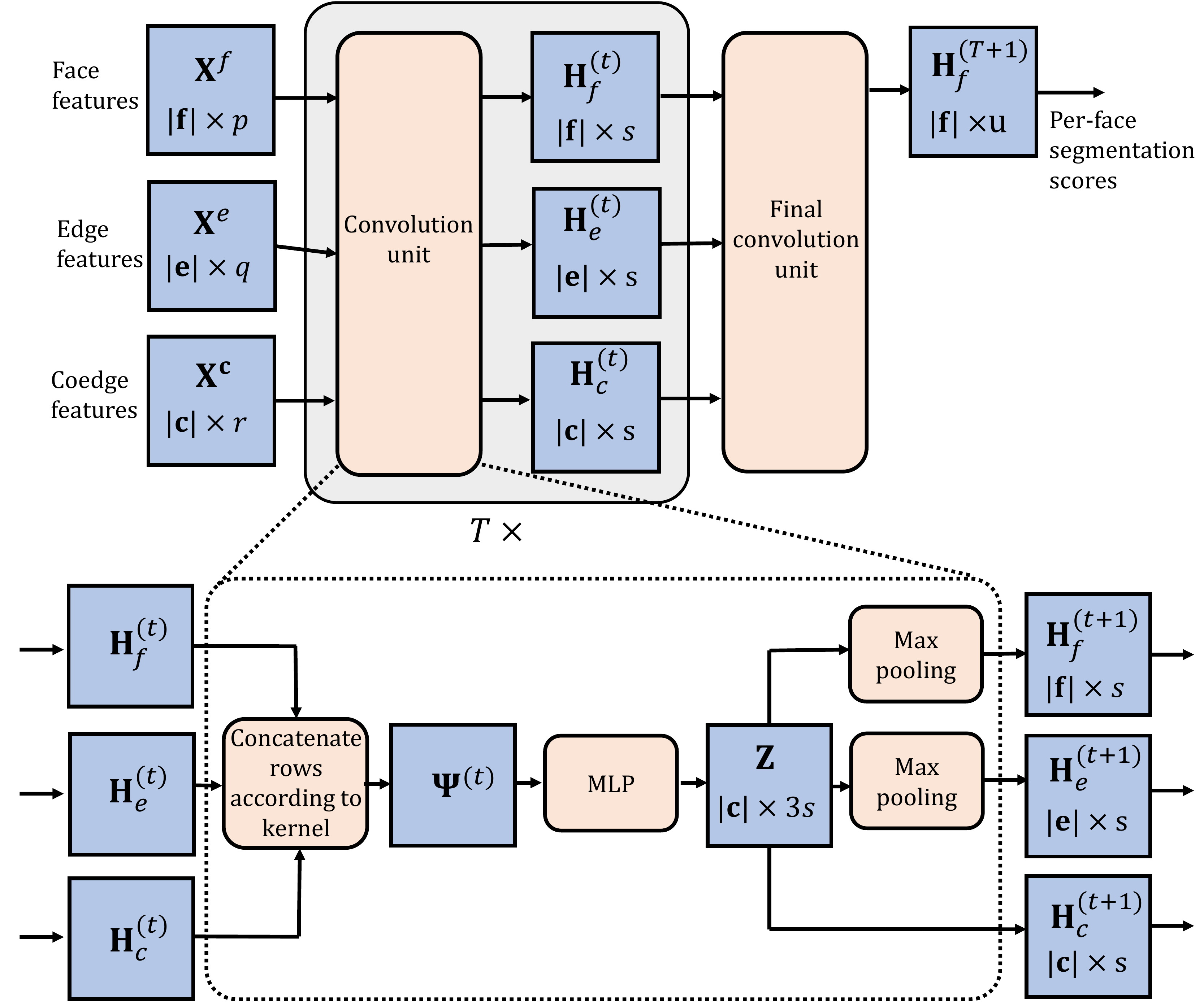}
    \caption{The BRepNet network architecture. Input feature vectors from faces, edges and coedges are passed through a stack of $T$ convolution units to generate hidden states ${\mathbf{H^{(t)}_f}}$, ${\mathbf{H^{(t)}_e}}$ and ${\mathbf{H^{(t)}_c}}$. A final convolution unit generates only the segmentation scores for faces.}
    \label{figure:convolution_unit}
\end{figure}

A forward pass through the network proceeds as follows.   We start by initialising the matrices ${\mathbf{H^{(0)}_f} = \mathbf{X^f}}$, ${\mathbf{H^{(0)}_e} = \mathbf{X^e}}$ and ${\mathbf{H^{(0)}_c} = \mathbf{X^c}}$.  These three matrices are then passed through a number of the convolution units as shown in Figure \ref{figure:convolution_unit}.  Following convolution unit $t$, the hidden state matrices  ${\mathbf{H^{(t)}_f}}$, ${\mathbf{H^{(t)}_e}}$ and ${\mathbf{H^{(t)}_c}}$ are generated.   The width of these hidden states is defined by a hyper-parameter $s$.  For face classification tasks a final convolution unit generates only matrix ${\mathbf{H^{(T+1)}_f} \in \R ^{|\mathbf{f}| \times u}}$ which are the per-face segmentation scores for each of the $u$ classes.

Inside each convolution unit three processes take place.  First we build up a matrix $\mathbf{\Psi}$ where
\begin{equation}
\begin{aligned}
    &\mathbf{\Psi^f} =  \bigparallel_{i=1}^{|K^f|} K^f_i \mathbf{H^{(t)}_f} \qquad
    &\mathbf{\Psi^e} =  \bigparallel_{i=1}^{|K^e|} K^e_i \mathbf{H^{(t)}_e} \\
    &\mathbf{\Psi^c} =  \bigparallel_{i=1}^{|K^c|} K^c_i \mathbf{H^{(t)}_c} \qquad
    &\mathbf{\Psi} = \mathbf{\Psi^f} || \mathbf{\Psi^e}  || \mathbf{\Psi^c}
\end{aligned}
\label{eq:fetch_and_concat}
\end{equation}
This procedure populates the $i$th row of $\mathbf{\Psi}$ with the concatenated hidden state vectors of the entities defined by the kernel with starting coedge $c_i$.

Each row of $\mathbf{\Psi}$ is then passed through a multi-layer perceptron (MLP) with parameters $\Theta^{(t)}$ and ReLU non-linearities.  The input to the first layer of the MLP depends on the number of columns of $\mathbf{\Psi}$ while all other MLP layers have a size 3 times the width of the hidden states $s$.

\begin{figure*}[!t]
\begin{center}
  \includegraphics[width=1\linewidth]{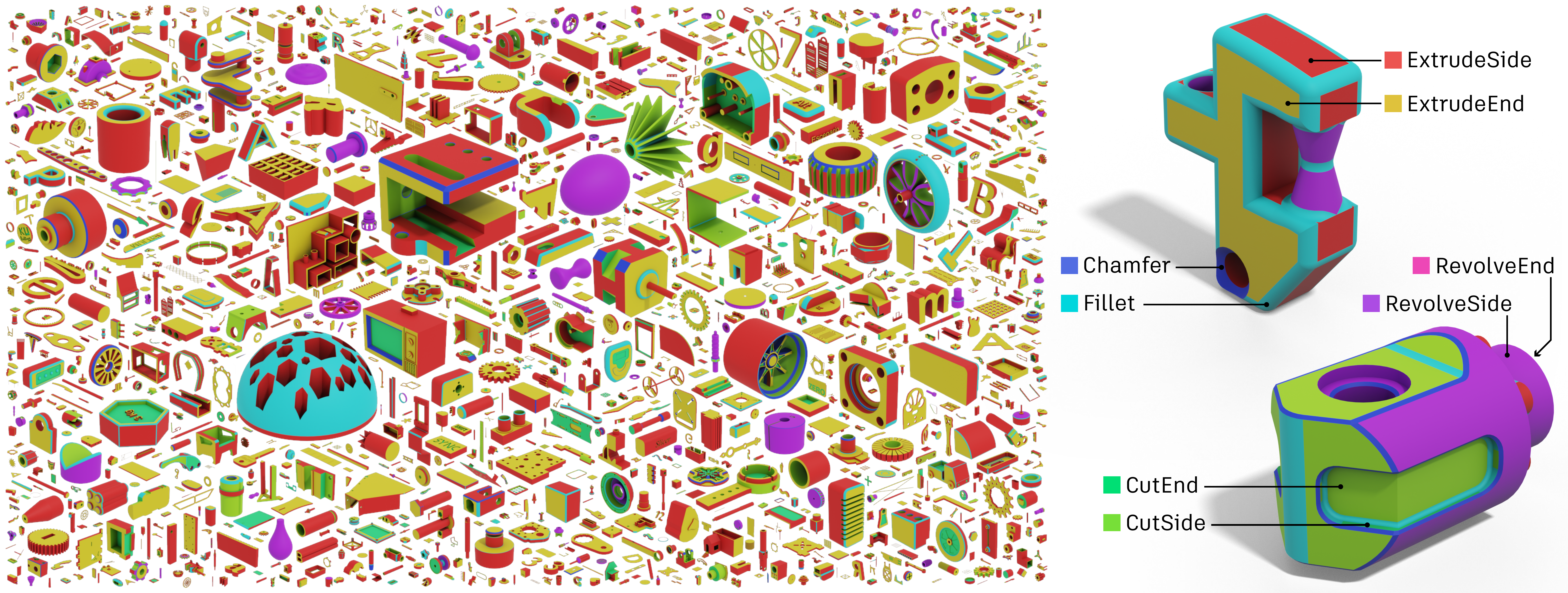}
\end{center}
   \caption{An overview of 3D models from the Fusion 360 Gallery segmentation dataset (left). Each 3D model is labeled according to the CAD modeling operations used to create it (right). }
   
\label{fig:data-overview}
\end{figure*}
Following the MLP, we generate a matrix $\mathbf{Z} \in \R^{|\mathbf{c}| \times 3s}$. The rows of $\mathbf{Z}$ are associated with coedges in the B-rep.   A simple architecture would pass the single matrix $\mathbf{Z}$ to the subsequent convolution units, however we observe that this simple approach gives poor performance on B-rep models where faces have multiple loops (e.g. a face with a hole).  In this case the edges of the B-rep do not form a connected graph and information cannot flow between the loops of multi-loop faces.  The performance of the network is greatly enhanced by pooling information from the coedges onto their parent faces and edges in each convolution unit.  This allows information to flow from the coedges in one loop onto the parent face, making it accessible to coedges in another loop in subsequent layers. To apply this pooling the matrix $\mathbf{Z}$ is first split into 3 sub-matrices of size $|\mathbf{c}| \ \times s$.

\begin{equation}
    \mathbf{Z} = \left[\begin{matrix} \mathbf{H_c^{(t+1)}} & \mathbf{Z^f} & \mathbf{Z^e} \end{matrix} \right]
\end{equation}
$\mathbf{H_c^{(t+1)}}$ is the matrix of hidden states for the coedges in the next layer, which requires no further processing. To build the $i$th row of the matrix $\mathbf{H_f^{(t+1)}}$ we apply element wise max pooling over the rows of $\mathbf{Z^f}$ corresponding to the coedges with parent face $f_i$.  The matrix $\mathbf{H_e^{(t+1)}}$ is built in a similar way by max pooling the pairs of rows of $\mathbf{Z^e}$ corresponding to coedges with the same parent edge.  
 
A diagram showing the matrices and operations performed in each convolution unit are shown in Figure \ref{figure:convolution_unit}.

\subsection{Face classifications}

The per-face segmentation scores for each class $u_i$ can then be calculated as follows.  In the final convolution unit, the last layer of the MLP has just $|u|$ neurons and produces only the matrix $\mathbf{Z}^f \in \R^{|\mathbf{c}| \times |u|}$.  The matrix of segmentation scores, $\mathbf{H}^{(T+1)}_f \in \R ^{|\mathbf{f}| \times |u|}$, is then built by pooling the coedge feature vectors onto their parent faces as before.  A cross-entropy loss can then be used to train the network.

%% file: sec/data.tex
\section{Fusion 360 Gallery segmentation dataset}
\label{sec:data}
In this section we introduce, to our knowledge for the first time, a dataset containing segmentation information for B-rep models and the corresponding triangle meshes and point clouds. The \textit{Fusion 360 Gallery} segmentation dataset is produced from designs submitted by users of the CAD package Autodesk Fusion 360 and is segmented based on the CAD modeling operations used to create each face. This modeling history information is not available in existing datasets \cite{koch2019abc, zhang2018featurenet, FabWave2019} and goes beyond \textit{what} was designed, providing insights into \textit{how} people design 3D models. 

The segmentation dataset contains a total of 35,858 3D models with per-face, per-triangle, and per-point segment labels provided for the B-rep, mesh and point cloud representations (Figure~\ref{fig:data-overview}, left). For segmentation we use a small subset of the most common CAD modeling operations: \textit{extrude}, \textit{chamfer}, \textit{fillet}, and \textit{revolve}.
In order to create a segmentation which contains as much information as possible about the CAD modeling operations, we subdivide extrude operations into additive (i.e. adding) and subtractive (i.e. cutting) extrusion operations, and further divide the faces created by extrude and revolve into side and end faces. This gives a set of eight labels for each face: \textit{ExtrudeSide}, \textit{ExtrudeEnd}, \textit{CutSide}, \textit{CutEnd}, \textit{Fillet}, \textit{Chamfer}, \textit{RevolveSide}, and \textit{RevolveEnd} (Figure~\ref{fig:data-overview}, right). Further details on the dataset are provided in the supplementary material.

%% file: sec/experiments.tex
\begin{figure*}[!t]
    \begin{minipage}{0.50\linewidth}
        \includegraphics[width=\columnwidth]{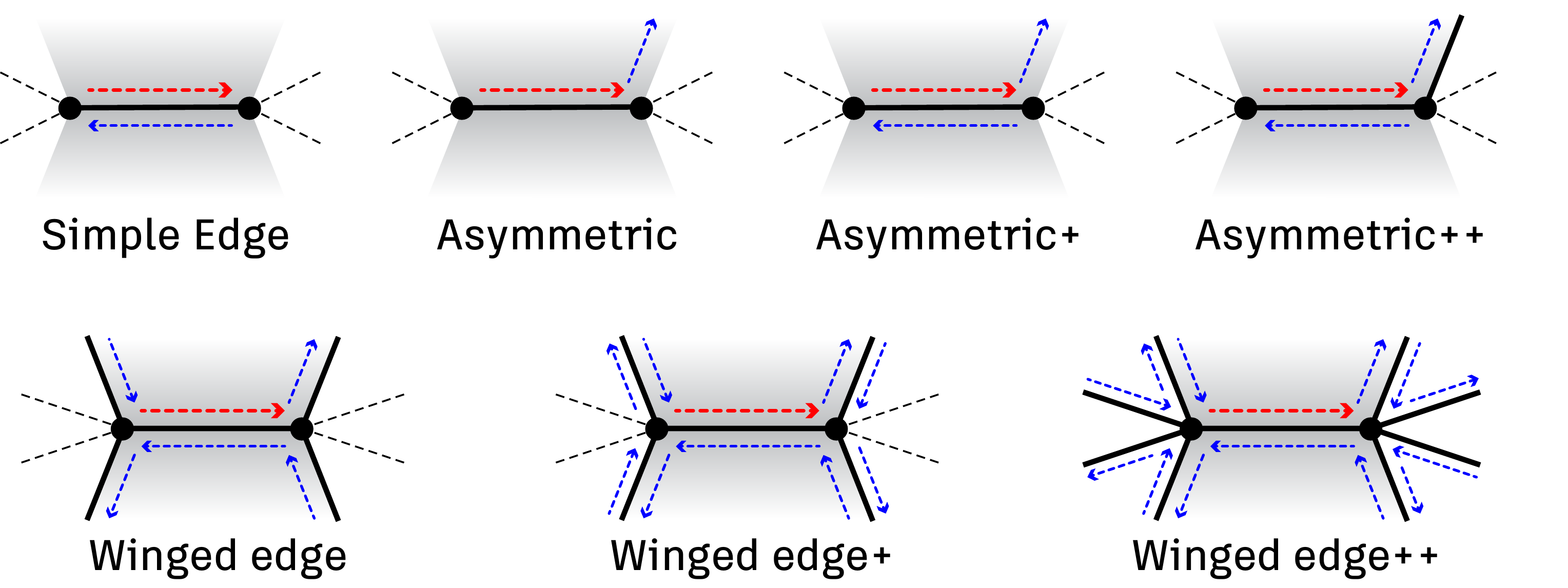}
        \par\vspace{0pt}
    \end{minipage}%
    \begin{minipage}{0.50\linewidth}
        \small
        \begin{tabular}{l||c|c|c|c}
            \textbf{Kernel} & $s$ &\textbf{$|\Theta|$} & \textbf{Accuracy \%} & \textbf{IoU \%} \\
            \hline
            Simple edge &  120 & 359k & 91.02 $\pm$ 0.20 & 74.03 $\pm$ 0.55\\
            Asymmetric &  120 & 359k & 91.66 $\pm$ 0.17 & 75.01 $\pm$ 0.54\\
            Asymmetric+ & 113 & 358k & 91.82 $\pm$ 0.13 & 75.06 $\pm$ 0.71\\
            Asymmetric++ & 107 & 359k & 92.05 $\pm$ 0.07 & 75.69 $\pm$ 0.38\\
            Winged edge & 84 & 359k & \textbf{92.52} $\pm$ 0.15 & 77.10 $\pm$ 0.54\\
            Winged edge+  & 75 & 357k & 92.50 $\pm$ 0.15 & 76.86 $\pm$ 0.47\\
            Winged edge++  & 63 & 358k & 92.50 $\pm$ 0.12 & \textbf{77.14} $\pm$ 0.44\\
            ECC & 153 & 360k & 90.36 $\pm$ 0.23 & 72.08 $\pm$ 0.50 \\
        \end{tabular}
        \par\vspace{0pt}
    \end{minipage}
    \caption{Different BRepNet kernel configurations (left) for which the accuracy and IoU are compared (right). The accuracy and IoU for the Edge-Conditioned Convolution (ECC) graph network \cite{Simonovsky2017} discussed in Section \ref{graph-benchmark} is also shown.
    MLP width $s$ is adjusted to keep the total number of parameters \textbf{$|\Theta|$} in the network to around 360k. }
    \label{figure:kernel_comparison}
\end{figure*}
\section{Experiments}
\label{experiments}
In this section we perform experiments to examine the following important network capabilities.  First we show that loop ordering information is useful for solving a B-rep segmentation problem. We study how performance is affected when the incidence relations in the matrices $\mathbf{N}$ and $\mathbf{P}$ are withheld from our architecture and explore a range of kernel configurations to find which one is optimal.  
We analyze the features passed to the network, identify the key information used to generate the segmentation and provide insights on why these features are important.  To demonstrate the advantages of learning based approaches we compare with a traditional rule-based feature recognition algorithm.   Next, we compare BRepNet performance against an Edge-Conditioned Convolution (ECC) graph network \cite{Simonovsky2017, gilmer2017neural}.  This architecture is chosen as it can ingest the same input features as BRepNet, but employs no special techniques to exploit the manifold nature of the B-rep.  As such this comparison shows the gains which can be made when specific kernel weights operate on specific neighbouring nodes.

Finally, as B-rep models can be converted to meshes and point clouds we compare against networks using these representations.  We investigate the advantages of working directly with the B-rep data structure and challenges of using approximations to the true geometry.

The data is divided into a 70/15/15\% train/validation/test split.  In each of the experiments above the networks are trained for 50 epochs and the weights with the smallest validation loss are recorded.  The performance of these trained models on the test set is then evaluated. The reported values are the average over 10 runs with different random seeds and the error bars are computed as the standard deviation.
\subsection{Evaluation metrics}
\label{evaluation-metrics}
We use accuracy and intersection over union (IoU) to evaluate network performance.  Due to data imbalance in the \textit{Fusion 360 Gallery} segmentation dataset, the IoU metric is useful for providing insight into the performance on the rarer classes. Rather than computing IoU values for individual B-rep models and then averaging as in \cite{qi2017pointnet}, we consider the entire collection of all faces in the test set at once.  This methodology is referred to as ``part IoU" in \cite{le2019going} and avoids the special case when a B-rep model has no faces, either predicted or in the ground truth, for a given class.

\subsection{Choice of kernel}
\label{choice_of_kernels}

The BRepNet architecture provides a flexible framework for defining the relative topological locations of the entities which make up a convolutional kernel.  Here we study how the choice of these entities affects network performance.  Figure \ref{figure:kernel_comparison}, left shows the range of different kernel configurations used in the experiments.  The corresponding topological walks are included in the supplementary material.   As the number of parameters in the MLP is dependent on the number of entities in the kernel, we adjust the hyper-parameter $s$ to keep the total number of network parameters as close as possible to 360k.   This decouples the effect of aggregating information from a wider region of the B-rep and the effects of increasing network capacity.  For each kernel configuration we train a network with two convolutional units, each with a two layer MLP.  Figure \ref{figure:kernel_comparison}, right shows the accuracy and IoU for each kernel configuration along with the values of $s$ and the corresponding number of parameters.

The ability of the network to exploit loop ordering information can now be evaluated. The ``simple edge" and ``asymmetric" kernels are carefully chosen to have the same number of faces, edges and coedges, allowing them to be compared directly without any adjustments in the MLP width.   The ``simple edge" arrangement contains only an edge and its two adjacent faces and coedges, giving it information similar to a face adjacency graph, but withholding information regarding the order in which coedges are arranged around the loop.   The ``asymmetric" kernel includes the next coedge in the loop in place of the mating coedge, allowing the kernel to observe patterns like contiguous smooth edges.   We observe $0.98\%$ improvement in IoU when moving from the ``simple edge" to ``asymmetric" kernels.  While this improvement is less than 2 standard deviations, a Welch's unequal variances t-test  gives a $P$ value of $0.0012$ for this result, indicating that the coedge ordering information is useful for the segmentation task.  

The ``winged edge" kernel configuration is similar to the \emph{half-flaps} described by Liu \etal \cite{Liu2020}.  It achieves an accuracy of $92.52\%$ and an IoU of $77.10\%$, over $5$ standard deviations above the IoU value achieved by the ``simple edge" kernel.  Adding additional entities to the kernel results in only very marginal gains as shown in the table at the right of  Figure \ref{figure:kernel_comparison}. This can be understood intuitively as the ``winged edge" kernel includes a compact set of topological entities immediately adjacent to a given edge.  When the kernel is expanded beyond this size, the locations at which the topological walks terminate become dependent on the B-rep topology in the vicinity of the edge.  For example the ``winged edge++" kernel configuration includes walks like $\mathbf{NMN}$ and $\mathbf{MPM}$ which will evaluate to the same entity when walking around vertices of valance 3 but distinct entities when the vertex has valance 4 or higher.  The ``winged edge" kernel lies at a sweet spot containing enough entities to allows patterns in local regions of the B-rep topology to be recognized, while being small enough not to be adversely affected by differences in the topology.

\subsection{Ablation studies on input features}
\label{ablation-studies}

\begin{figure}[t]
    \includegraphics[width=\columnwidth]{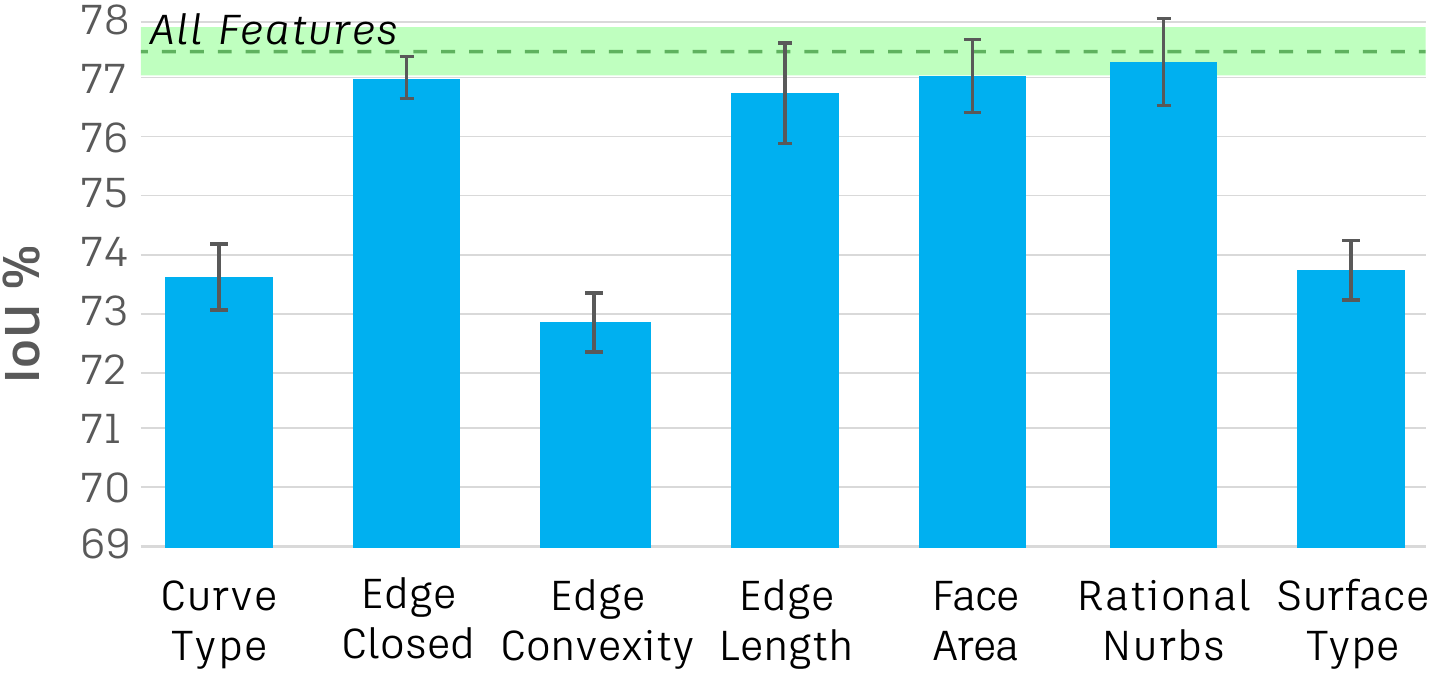}
    \caption{Effect on IoU of removing groups of input features from the network.}
    \label{fig:ablation}
\end{figure}

Here we identify which of the input features described in Section \ref{section:input_features} play an important role in the results for the segmentation.   The network is trained with groups of input features removed and the resulting IoU values are shown in Figure \ref{fig:ablation}.  The ``winged edge+" kernel configuration is employed in these experiments and the hyper parameters are as described in Section \ref{choice_of_kernels}.  We see that removing the one-hot encodings for surface type reduces IoU by $3.7\%$ and removing curve type information reduces the IoU by $3.9\%$.  These large reductions in performance are expected as the surface and curve type information is the primary way geometric information is fed to the network.  

We also observe a $4.6\%$ reduction in IoU when edge convexity is removed.  Edge convexity is well known to be useful for the detection of form features and was used in a large number of early neural networks \cite{Prabhakar1992, nezis1997, Sunil2009, Shi2020}.  Joshi \etal \cite{Joshi1988GraphbasedHF} offers an insight into how edge convexity could be useful with the observation that a face with all convex edges cannot be a part of a concave feature (\textit{CutSide} or \textit{CutEnd}).

Removing other input features have much smaller effects.  Without the edge length feature the IoU only decreases by $0.7\%$ and removing the face area feature causes just a $0.4\%$ IoU decrease.  Hence we conclude that edge convexity, curve type and surface type are the primary pieces of information used by the network in segmentation.

\begin{table*}
    \centering
    \small
    \begin{tabular}{c||c|c|c|c|c}
         & \textbf{Per-face accuracy}  & \textbf{Per-face IoU} & \textbf{Per-entity accuracy} & \textbf{Per-entity IoU} & \textbf{$|\Theta|$}\\
         \hline
         BRepNet & \textbf{92.52 $\pm$ 0.15} & \textbf{77.10 $\pm$ 0.54} & - & - & \textbf{359,100} \\
         ECC &  90.36 $\pm$ 0.23 & 72.08 $\pm$ 0.50  & - & - & 359,558 \\
         ASM & 64.57 & 49.53 & - & - & - \\
        PointNet++ & 74.00	$\pm$ 0.84 & 34.78 $\pm$ 2.23 & 82.78 $\pm$ 0.30 & 36.49 $\pm$ 1.67 & 1,403,784 \\
         MeshCNN & 62.99 $\pm$ 0.37 & 20.59 $\pm$ 0.41 & 73.81 $\pm$ 0.51 & 24.07 $\pm$ 0.42 & 2,279,720 \\
    \end{tabular}
    \caption{Accuracy, IoU and number of model parameters, \textbf{$|\Theta|$}, for a variety of different networks.  The BRepNet results are for the ``winged edge" kernel configuration. The per-entity accuracy and IoU columns refer to per-edge accuracy in the case of MeshCNN and per-point accuracy in the case of PointNet++. }
    \label{tab:benchmark}
\end{table*}

\subsection{Heuristic method comparison}

Many CAD modeling packages use rule-based algorithms for the detection of form features.  Here we compare against the feature recognition capabilities of Autodesk Shape Manager (ASM)~\cite{asm}, an industry standard CAD kernel used in numerous commercial products. 

As ASM cannot detect the \textit{RevolveEnd} segment type, we omit this when computing the ASM average IoU result.  ASM does not identify any modeling feature type for 13\% of the faces in the dataset and we consider these faces to be incorrectly classified.  

The results for ASM feature recognition on the \textit{Fusion 360 Gallery} segmentation task are shown in Table \ref{tab:benchmark}. While BRepNet achieves an IoU value 27\% higher than ASM, this does not reflect the results qualitatively.  When features are recognized by the ASM algorithm, the faces identified are always geometrically consistent with the type of feature found.  The confusion is between classes where the actual modeling technique used is ambiguous.  For example, a designer may choose to create a cylinder with an extruded circle or a revolved rectangle.
The higher accuracy achieved on the classification task by BRepNet shows the network is capable of learning the \textit{most likely} modeling technique a designer will employ rather than simply identifying one of the possible solutions.     

\subsection{Edge-conditioned convolution graph network}
\label{graph-benchmark}
In this section we compare BRepNet performance with an Edge-Conditioned Convolution (ECC) graph network as described in \cite{Simonovsky2017}.  As discussed in Section \ref{ablation-studies}, an important indicator for the class of a face is the convexity of its surrounding edges.  As this architecture allows edge attributes to affect the messages passed between faces it is well suited to \textit{Fusion 360 Gallery} segmentation task.
The B-rep topology is translated into a face adjacency graph with the faces represented as nodes connected by pairs of directed arcs with opposite orientations.  We use the face features $\mathbf{X^f}$ as input node features.   As the directed arcs map 1:1 with the coedges in the B-rep, we create the attribute vectors for each directed arc by concatenating the corresponding coedge features with the features of its parent B-rep edge.

We match the hyper-parameters of the network to those of BRepNet as closely as possible.  Two edge-conditioned convolution layers are used, with the edge specific weight matrices computed by two-layer MLPs.  The width of the first MLP input is defined by the number of face features and all subsequent widths were set to 153.  This gave the ECC network a total of $359,558$ parameters which is the closest possible match to BRepNet.
The accuracy and IoU of the ECC network are shown in Figure \ref{figure:kernel_comparison} and Table \ref{tab:benchmark}.  The IoU value is more than $5\%$ below what BRepNet can achieve using the `winged edge" kernel.  This is expected as this graph network architecture is not specifically designed for convolution on manifolds and does not map specific learnable parameters to specific entities in the convolution.  We would expect to see IoU values approximately equal to what BRepNet achieves with the ``simple edge" kernel, but on the test set the ECC gives a $2\%$ lower IoU value.  We noticed that BRepNet is more stable than the ECC during training, and we believe the poor performance of the ECC on the test set may be partially due to the choice of epoch for which the trained model was recorded for use at test time.  In all experiments the model with the lowest validation loss is used for evaluation on the test set.  

\subsection{Comparison with geometry based methods}
\label{experiments-benchmark}

In this section we investigate the advantages of working directly with B-rep models rather than converting the geometry to meshes or point clouds.  We generate closed and manifold meshes from the B-rep geometry with close to 3000 triangles edges each. Computing meshes which meet this criteria is itself a difficult task requiring specialized meshing algorithms.  For $13\%$ of B-rep models the meshing algorithm failed entirely and the B-rep representations of these models were removed from the dataset.  Avoiding the requirement to generate these high quality meshes is a major advantage of working directly on B-rep data.

Point cloud data is then generated by random uniform sampling of 2048 points over the surface of each mesh. As the number of points sampled from each face is determined by area, small faces generate a low number of points.  The potential to under-sample some faces is a disadvantage for point cloud techniques as small holes and grooves, which can be critical for the function of the part, may be missed. 

Two well-known architectures, PointNet++~\cite{qi2017pointnet++}, and MeshCNN~\cite{Hanocka2019} are adapted for the face segmentation task. To compare performance between multiple representations we use per-face classification accuracy and IoU as our primary evaluation metrics. The triangles generated from each B-rep face inherit the label of that face and points inherit the labels of the triangles from which they were sampled. Triangle edges are considered to be owned by the first of the two triangles sharing the edge and edge labels are derived from the faces which generated the triangles. The per-face accuracy and IoU is then evaluated by averaging the segmentation scores for the points or edges derived from each face. This gives a prediction of the class for each face, from which the accuracy and mean IoU can be evaluated as described in Section \ref{evaluation-metrics}. In addition to the per-face metrics we also report the per-point and per-edge accuracy and IoU for PointNet++ and MeshCNN respectively.  As for the per-face metrics, the IoU values are computed by considering the points or edges from all bodies together. 

Table~\ref{tab:benchmark} details the accuracy and IoU results of the segmentation task along with the number of model parameters. Both BRepNet and the ECC easily outperform the geometry based methods by more than $16\%$ accuracy and $37\%$ in IoU with just under $1/4$ the number of parameters.  This demonstrates the advantages of working directly with the compact B-rep data rather than derived representations.  

%% file: sec/conclusions.tex
\section{Conclusions}
We have presented BRepNet, a neural network architecture which can operate directly on B-rep models. We also introduced the \textit{Fusion 360 Gallery} segmentation dataset and provide benchmark results on the segmentation problem. Our results show that by using the concise surface and edge type information from the B-rep data structure the network can easily outperform existing techniques using point clouds and meshes on the \textit{Fusion 360 Gallery} dataset segmentation task.  In addition we demonstrate that by defining convolutional kernels relative to the coedges of the B-rep, the architecture can make use of information about the next and previous coedges in the loops around faces, giving better performance than an edge conditioned convolution network with the same number of parameters.  In future work we plan to apply this convolution scheme to other problems where graphs are embedded in 2D manifolds such as polyhedral models, subdivision surfaces, super-pixel segmentations of image data, region growing algorithms on triangle meshes and for learning tasks on Voronoi diagrams.

%% file: sec/appendix.tex
\clearpage
\setcounter{section}{0}
\renewcommand\thesection{\Alph{section}}
\renewcommand\thesubsection{\thesection.\arabic{subsection}}

\section{Supplementary Material}
\input{sec/appendix_dataset}
\input{sec/appendix_kernels}
\input{sec/appendix_experiments}

%% file: sec/appendix_dataset.tex
\subsection{Dataset statistics}
\label{sec:datastats}
\begin{figure*}
    \centering
    \includegraphics[width=0.95\textwidth]{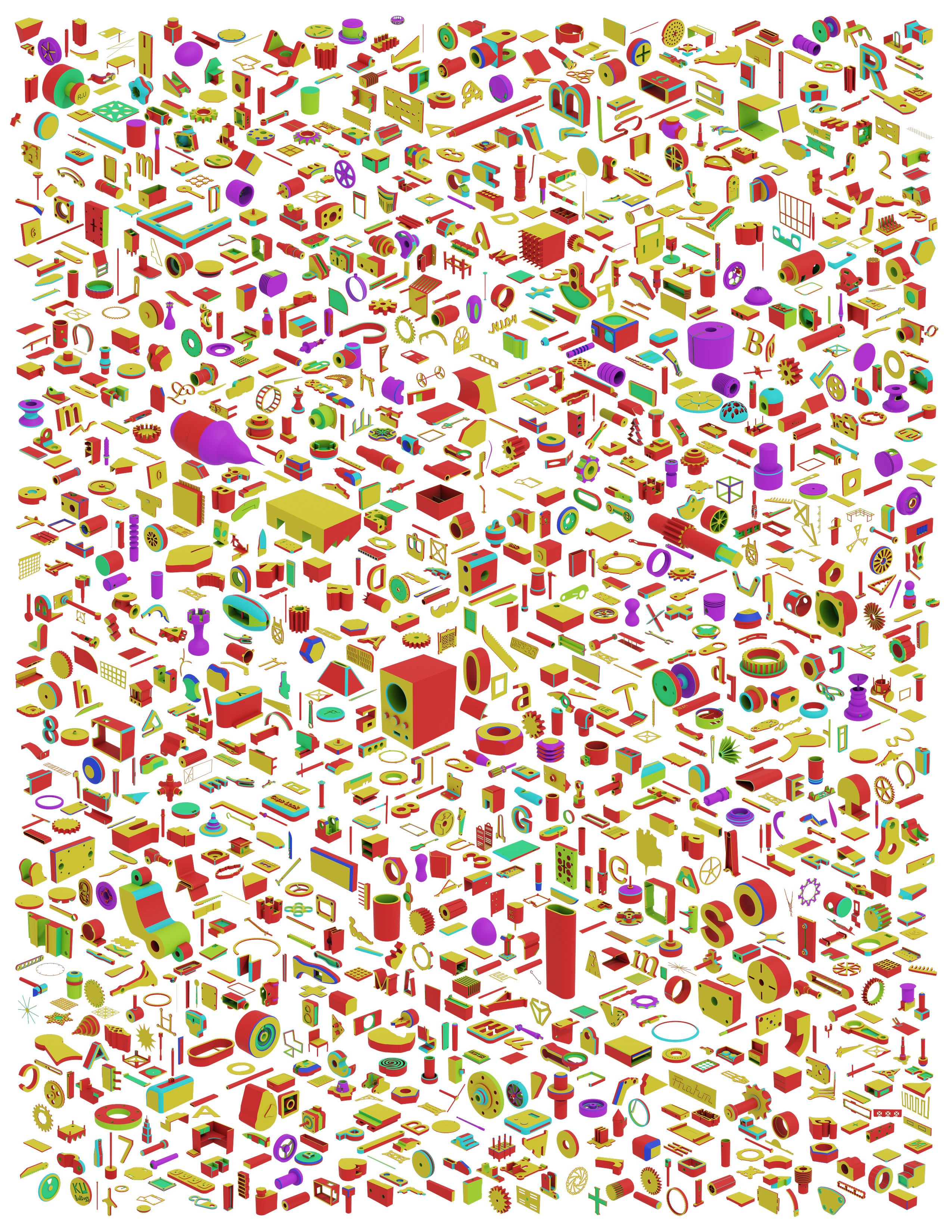}
    \caption{ Example designs from the \textit{Fusion 360 Gallery} segmentation dataset, colored by segmentation label. }
    \label{fig:dataset_mosaic_appendix}
\end{figure*}

The  \textit{Fusion 360 Gallery} segmentation dataset contains a total of  35,858 B-rep bodies with corresponding segmentation information, high quality triangle meshes and point clouds.   The train/test split used in this work is published with the dataset.   It contains $5,399$ bodies in the test set, with the remaining $30,459$ bodies for use training and performing validation. Example designs from the dataset are shown in Figure~\ref{fig:dataset_mosaic_appendix}, colored according to segmentation label.

\begin{figure}
     \includegraphics[width=1\linewidth]{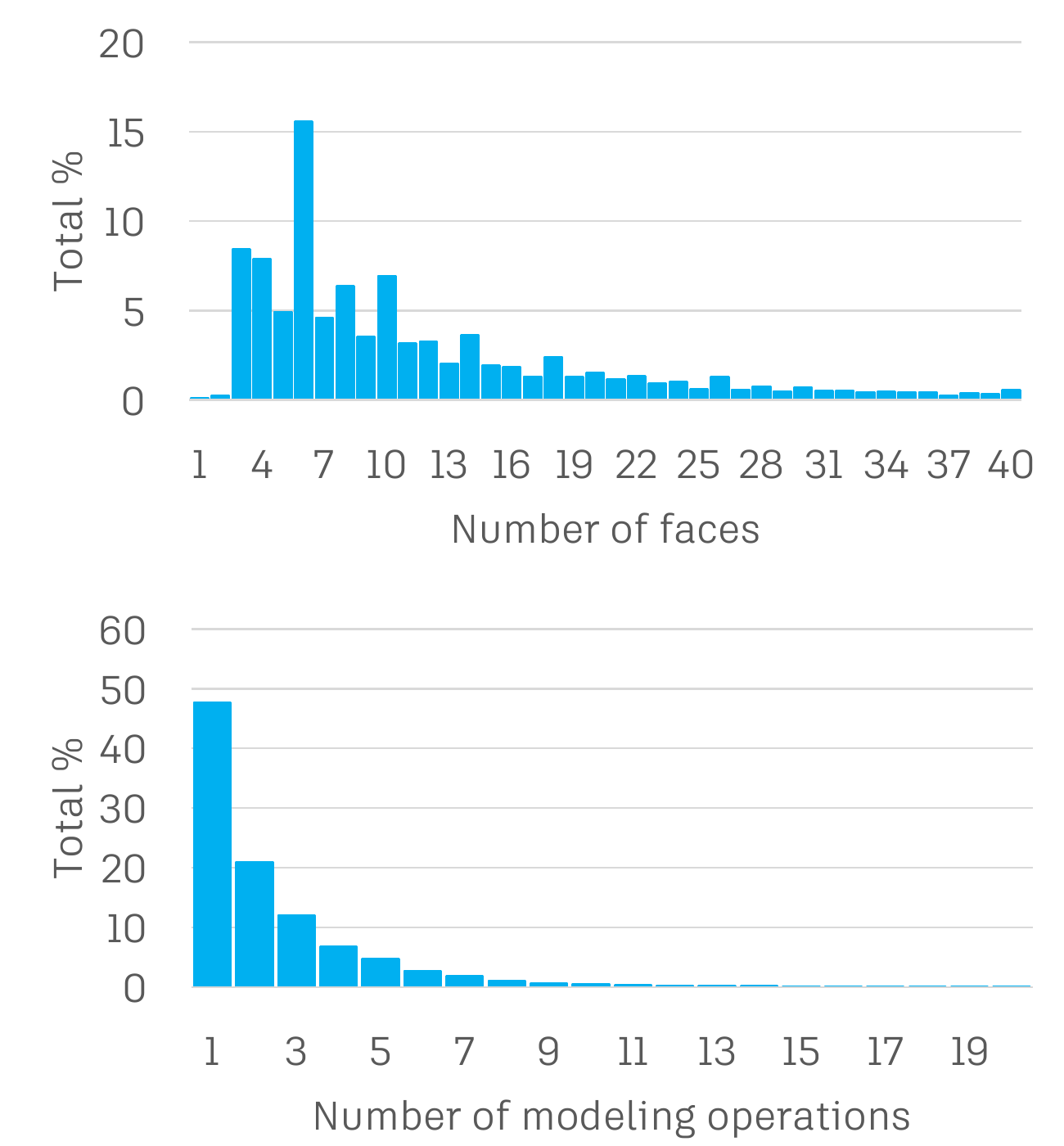}
     \caption{The distribution of faces per body (top) and of CAD modeling features used to generate each body (bottom).}
     \label{fig:body_stats}
\end{figure}

The complexity of the models can be understood from the number of faces per B-rep body and the number of CAD modeling operations used in their construction.  Histograms of these distributions are shown in Figure \ref{fig:body_stats}.  Many of the bodies are relatively simple with half of them having fewer than $9$ faces.  There is a long tail of more complicated B-reps with the most complex having 421 faces.   Just over half of the bodies were constructed with more than one CAD modeling operation and $31\%$ have two or more.  Only $1\%$ of the bodies used more than $10$ operations in the construction history and the maximum number of operations used to create any B-rep in the dataset is 59.

\begin{figure}[!t]
     \includegraphics[width=1\linewidth]{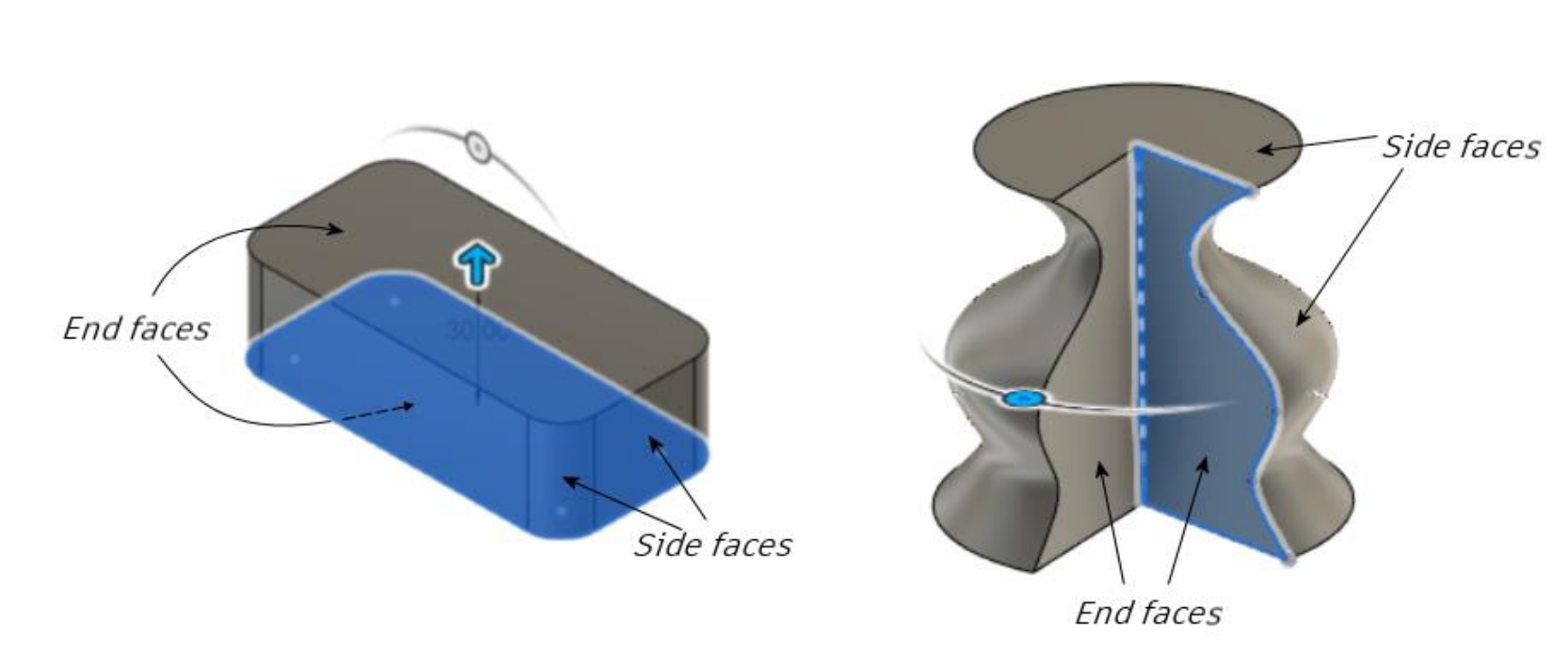}
     \caption{Side and end faces for extrusions and revolutions.}
     \label{fig:label_start_end}
\end{figure}

As explained in Section \suppref{sec:data}{4}, the dataset is modeled entirely using extrusions, revolutions, fillets and chamfers.  The Autodesk Fusion 360 CAD package allows the  geometry/topology created by each modeling operation to be immediately combined with the body being constructed using either a boolean union, subtraction or intersection.  Extrusions are the most common way geometry is created, comprising $74\%$ of modeling operations.   To avoid a very large imbalance in the dataset, the collection of faces created by extrusions are subdivided as follows.   When the extrusions were used to create new bodies or unioned with to existing bodies, the faces are placed in the \emph{Extrude} class, while faces from extrusions which are subtracted from or intersected with existing bodies are placed in the \emph{Cut} class.   The faces generated by extrusions and revolutions are then further subdivided into side and end faces.  For extrusions, side faces are created by sweeping the profile geometry while end faces are parallel to the plane on which the profile was sketched.  For revolutions, sides faces are created by revolving the profile.   \textit{RevolveEnd} faces are only created when the extrusion is not 360 degrees as shown in Figure \ref{fig:label_start_end}. This results in eight possible labels: \textit{ExtrudeSide}, \textit{ExtrudeEnd}, \textit{CutSide}, \textit{CutEnd}, \textit{Fillet}, \textit{Chamfer}, \textit{RevolveSide}, and \textit{RevolveEnd}.

\begin{figure}[!t]
     \includegraphics[width=1\linewidth]{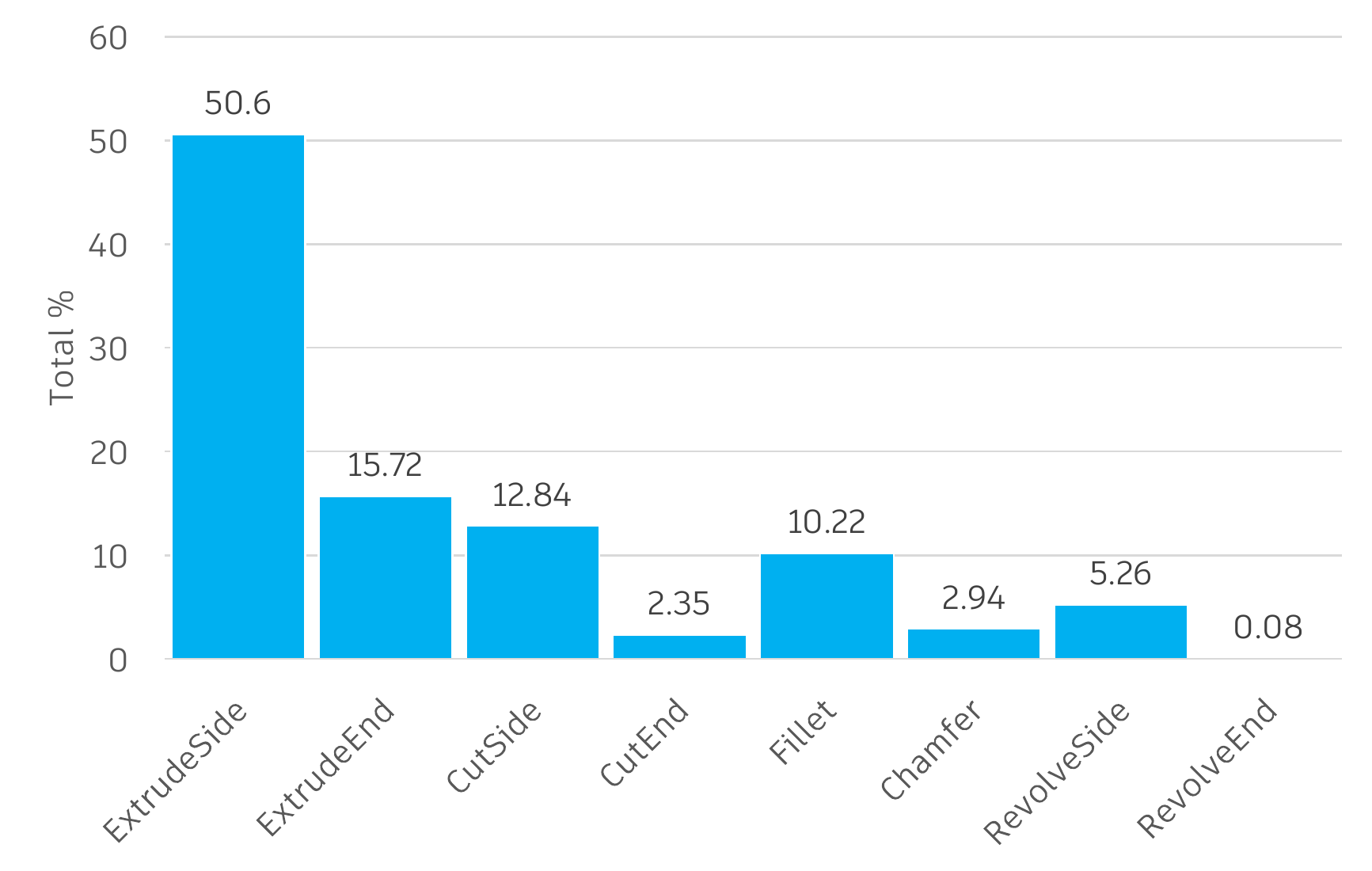}
     \caption{The percentage of faces in each class.}
     \label{fig:label_balance}
\end{figure}

\begin{table*}
    \begin{center}
    \begin{tabular}{l|l|p{0.25\linewidth}|p{0.25\linewidth}}
    \small
    \textbf{Kernel} & Faces &Edges & Coedges \\
    \hline
    Simple edge & 
                $\mathbf{F}$, $\mathbf{MF}$&
                
                $\mathbf{E}$ & 
                
                $\mathbf{I}$, $\mathbf{M}$ \\

    Asymmetric &    
            $\mathbf{F}$, $\mathbf{MF}$&
            
                $\mathbf{E}$ &
                
            $\mathbf{I}$, $\mathbf{N}$ \\

    Asymmetric+ &     
            $\mathbf{F}$, $\mathbf{MF}$&
            
            $\mathbf{E}$ &
            
            $\mathbf{I}$, $\mathbf{M}$, $\mathbf{N}$\\

    Asymmetric++ &   
            $\mathbf{F}$, $\mathbf{MF}$&
            
            $\mathbf{E}$, $\mathbf{NE}$ & 
            
            $\mathbf{I}$, $\mathbf{M}$, $\mathbf{N}$\\

    Winged edge &   
            $\mathbf{F}$, $\mathbf{MF}$&
            
            $\mathbf{E}$, $\mathbf{NE}$, $\mathbf{PE}$,  $\mathbf{MNE}$, $\mathbf{MPE}$&
            
            $\mathbf{I}$, $\mathbf{M}$, $\mathbf{N}$, $\mathbf{P}$, $\mathbf{MN}$,$\mathbf{MP}$\\

    Winged edge+  &   
            $\mathbf{F}$, $\mathbf{MF}$& 
            
            $\mathbf{E}$, $\mathbf{NE}$, $\mathbf{PE}$,  $\mathbf{MNE}$, $\mathbf{MPE}$&
            
            $\mathbf{I}$, $\mathbf{M}$, $\mathbf{N}$, $\mathbf{NM}$, $\mathbf{P}$, $\mathbf{PM}$, $\mathbf{MN}$, $\mathbf{MNM}$, $\mathbf{MP}$, $\mathbf{MPM}$\\

    Winged edge++  & 
            $\mathbf{F}$, $\mathbf{MF}$& 
            
            $\mathbf{E}$, $\mathbf{NE}$, $\mathbf{PE}$,  $\mathbf{MNE}$, $\mathbf{MPE}$, $\mathbf{NMNE}$, $\mathbf{PMPE}$, $\mathbf{MPMPE}$, $\mathbf{MNMNE}$&
            
            $\mathbf{I}$, $\mathbf{M}$, $\mathbf{N}$, $\mathbf{NM}$, $\mathbf{P}$, $\mathbf{PM}$, $\mathbf{MN}$, $\mathbf{MNM}$, $\mathbf{MP}$, $\mathbf{MPM}$, $\mathbf{NMN}$, $\mathbf{PMP}$, $\mathbf{MPMP}$, $\mathbf{MNMN}$\\
    \end{tabular}
    \end{center}
    \caption{The topological walks making up the kernels shown in Figure \suppref{figure:kernel_comparison}{5}.}
    \label{table:kernel_walks}
\end{table*}

The fraction of faces with each label type is shown in Figure \ref{fig:label_balance}.  We see that just over half the faces are in the \textit{ExtrudeSide} class, making the dataset relatively imbalanced.  In particular the \textit{RevolveEnd} class is very rare, accounting for just $0.08\%$ of faces.  The \textit{CutEnd} and \textit{Chamfer} classes are also rare at $2.35\%$ and $2.94\%$ of the faces respectively.  The faces in these classes are often planar, so the correct segment type can only be identified by considering the surrounding shape.  This makes the identification of these rare classes an extremely challenging task.    
\begin{figure*}[!t]
     \includegraphics[width=1\linewidth]{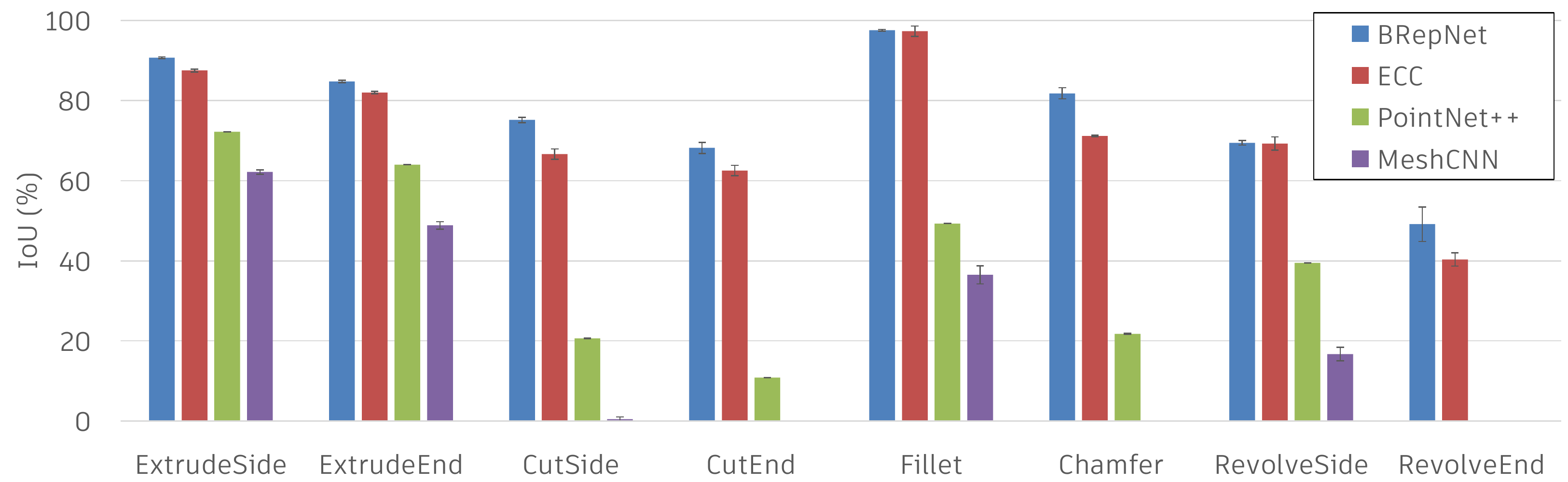}
     \caption{The per-face IoU values achieved by each network for each segment class.}
     \label{fig:IoU_per_class}
\end{figure*}
\subsubsection{Data preparation}
In this section we describe the processing performed on the designs from the Autodesk Online Gallery\footnote{https://gallery.autodesk.com/fusion360} to generate the \textit{Fusion 360 Gallery} segmentation dataset.  Scripts were used to drive the Application Programming Interface (API) of the  \textit{Fusion 360} CAD package to load each Fusion document and suppress all features except extrusions, revolutions, fillets and chamfers.  While this procedure modified the shapes of some models, it greatly increased the number of B-rep bodies available for the dataset.  

The number of faces and edges in the resulting bodies, along with the surface area and volume of each body, was then recorded.  To remove duplicate models, we first search for bodies with the same number of faces and edges.  The area and volume of these candidates are then checked and bodies for which the area and volume matches to within $1\%$ were discarded.  The de-duplication procedure was verified using thumbnails of the resulting duplicate free dataset.  These were ordered by surface area and images were manually inspected to verify that the area and volume tolerance was appropriate for detecting duplicate bodies, even when rigid body transforms had been applied.  Models with small topological differences were not considered to be duplicates as BRepNet is designed to be sensitive to differences in the model topology.

All the B-reps in the dataset were transformed to place the center of their bounding boxes at the origin.  A uniform scaling was then applied so that the largest length of the bounding box is $2$ model units across.  Meshes and point clouds  were then extracted from these scaled models.

\subsubsection{Input feature standardization}
The input features for BRepNet are standardized using the following procedure.  The mean, $\mu$ and standard deviation, $\sigma$, are computed for each input feature for the entire training set.  The standardized values $x'$ are then computed as

\begin{equation}
x'=\frac{x-\mu}{\sigma}
\end{equation}

\subsubsection{Support for operation grouping and ordering problems}
In addition to the segmentation data, we also provide more detailed information about the modeling operations used to construct the solid.  For each B-rep face we provide a unique identifier for the operation which created it, allowing groups of faces created by different operations of the same type to be separated.  For extrusions and revolutions we also provide a classification of the face as \emph{Start}, \emph{End} or \emph{Side}.  We provide the type of each operation and the order in which the operations were applied in the parametric model history.  For the point cloud and mesh representations we provide the mapping from each point and triangle to the face from which it was sampled, allowing this extra information to be used for all representations.   More details on how this data is organized is in the dataset documentation\footnote{https://github.com/AutodeskAILab/Fusion360GalleryDataset}.

This additional data is intended to support research into reverse engineering tasks.  Grouping the points or triangles according to the operation which created them is a first step towards rebuilding the parametric history.  Further subdividing each extrusion based on the \emph{Start}, \emph{End} and \emph{Side} information allows the extrusion direction to be predicted.  Slicing the mesh perpendicular to this extrusion direction can then allow the profile curves to be extracted and the extruded volumes regenerated.  Finally, by predicting the ground truth operation type and order, sensible sequences for the modeling operations can be learned.  

%% file: sec/appendix_kernels.tex
\subsection{Kernels}

In Section \suppref{choice_of_kernels}{5.2} the performance of BRepNet was compared with a number of different kernels.  Diagrams showing the entities taking part in these kernels are shown in Figure \suppref{figure:kernel_comparison}{5}.   In these diagrams each topological walk in the kernel terminates on a distinct entity.  It should be noted that this will not always be the case for arbitrary B-reps data.  Some local topology will result in two or more topological walks  terminating on the same entity.  
No special case handling is required when this happens.  The procedure described in Equation \suppref{eq:fetch_and_concat}{2} simply concatenates feature vectors from the same entity into the same row of $\Psi$ multiple times.  The network learns to recognizer these repeated patterns in the feature vectors in the same way as in the case where the entities are distinct.  Table \ref{table:kernel_walks} gives the full lists of topological walks which make up the kernels.

%% file: sec/appendix_experiments.tex
\begin{figure*}
    \centering
    \includegraphics[width=0.85\textwidth]{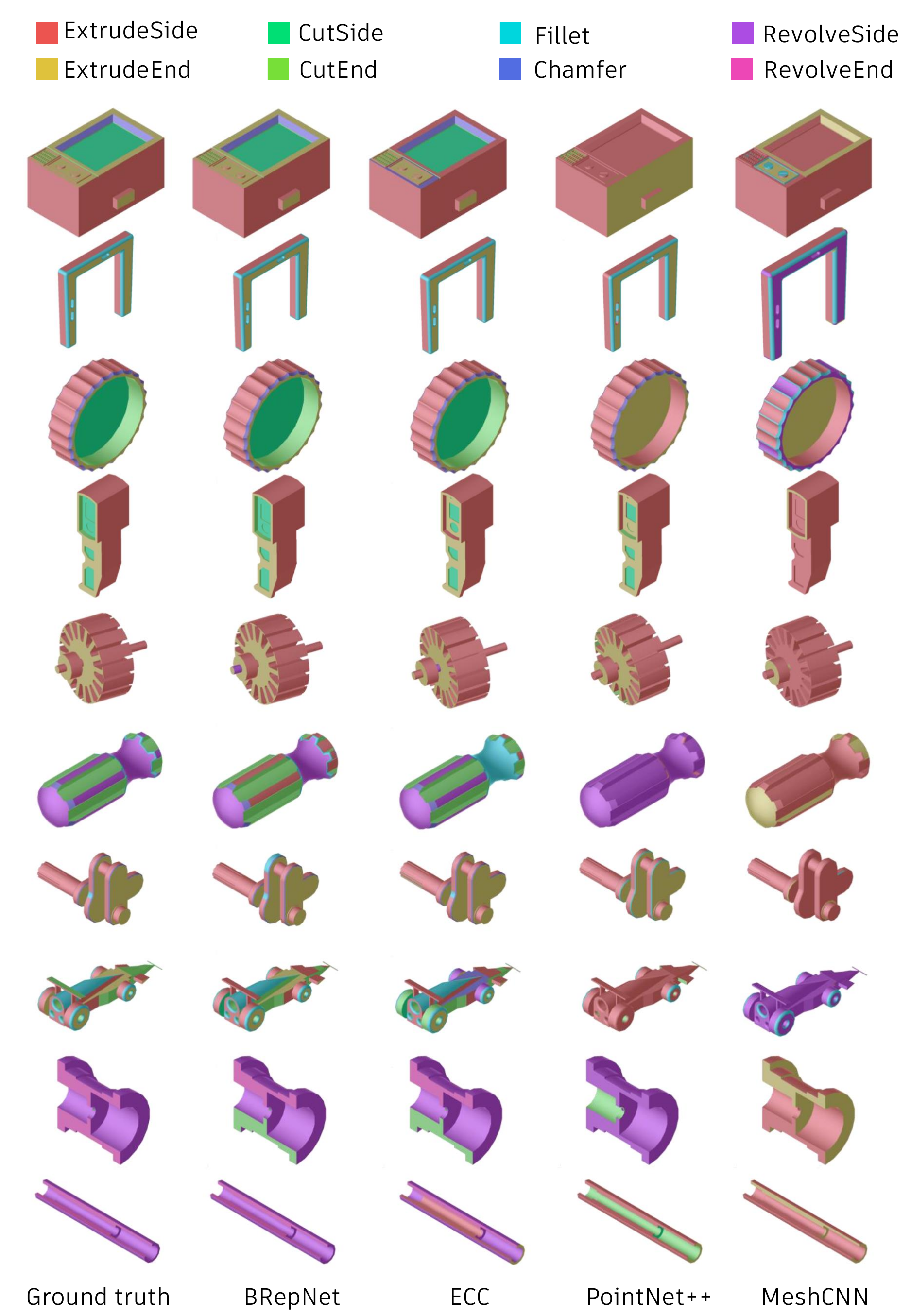}
    \caption{The per-face segmentation generated by each network.}
    \label{fig:result_images}
\end{figure*}
\subsection{Experiments}

\subsubsection{Training details}

For all the experiments described in Section \suppref{experiments}{5}, the BRepNet network was trained using the Adam optimizer, with default parameters (learning rate of $0.001$ and betas $0.9$ and $0.999$).   The B-reps in the dataset were divided into mini-batches, each containing approximately 1000 faces.   Multiple B-Reps can be combined into a same batch by row-wise concatenation of the input feature matrices $\mathbf{X}^f$, $\mathbf{X}^e$ and $\mathbf{X}^c$, and diagonal concatenation of the matrices  $\mathbf{N}$, $\mathbf{P}$, $\mathbf{M}$, $\mathbf{E}$ and $\mathbf{F}$.
Training and evaluate was performed using NVIDIA Tesla V100 GPUs.   Training took an average of 45s per-epoch with the network typically achieving a minimum loss value by around the $15$th epoch.  Hence the BRepNet network could be trained on the \textit{Fusion 360 Gallery} segmentation dataset in under 12 minutes from random seed.   

For experiments comparing against PointNet++ and MeshCNN we use the official implementation and retain the default hyper-parameters where possible. PointNet++ models are trained with a batch size of 32, a learning rate of 0.001 and momentum of 0.9 using the TensorFlow implementation\footnote{https://github.com/charlesq34/pointnet2} from~\cite{qi2017pointnet++}. The PyTorch implementation\footnote{https://github.com/ranahanocka/MeshCNN} of MeshCNN was used with a batch size of $12$, a learning rate of $0.0002$ and momentum of $0.9$.   The maximum number of input edges for any mesh was $3500$ and the pooling resolutions were set to $2500$,  $1750$ and $1000$. 

The ECC used the pytorch-geometric NNConv implementation\footnote{https://pytorch-geometric.readthedocs.io/en/latest/modules/nn.html}.   The Adam optimizer was used for the training with default parameters (learning rate of $0.001$ and betas $0.9$ and $0.999$).

\subsubsection{Comparison of IoU for different classes}
In this section we show the IoU values achieved by BRepNet, the Edge-Conditioned Convolution (ECC) graph network \cite{Simonovsky2017}, PointNet++ \cite{qi2017pointnet++} and MeshCNN \cite{Hanocka2019} for the different classes individually.  Figure \ref{fig:IoU_per_class} shows the per-face IoU values each network achieved on each class and Figure \ref{fig:result_images} shows images of the face segmentation on some example models.    We see that a key reason why BRepNet and the ECC network perform better than PointNet++ and MeshCNN is their ability to correctly classify faces in the rare classes.  

The \textit{RevolveEnd} class always consists of planar faces and accounts for just $0.08\%$ of the dataset.  While BRepNet finds this class challenging, achieving only $49\%$ IoU, both PointNet++ and MeshCNN fail to identify any  \textit{RevolveEnd} faces.  We believe this is because the \textit{RevolveEnd} class can only be identified by considering a face in the context of its neighbouring faces and edges.  This is something which the ECC and BRepNet approaches can do easily, as both networks are designed to leverage information from adjacent faces.  PointNet++ and MeshCNN have a hierarchical design which is intended to improve the flow of information from neighbouring geometry.  PointNet++ achieves an IoU of $39\%$ of the neighbouring  \textit{RevolveSide} faces, while MeshCNN achieves only $17\%$.  Neither architecture manages the extremely challenging task of using their identification of the \textit{RevolveSide} faces to correctly classify the adjacent groups  of planar points/edges as  \textit{RevolveEnd}.  

The \textit{CutEnd} class is also rare, accounting for just $2.35\%$ of faces in the dataset.  As for \textit{RevolveEnd}, these faces are always planar.  The primary way they can be distinguished from the more common \textit{ExtrudeEnd} faces is by observing that they are often surrounded by concave edges.  PointNet++ is able to correctly identify $10.83\%$ of \textit{CutEnd} faces, while MeshCNN recognizes just $0.01\%$ of them.  While MeshCNN uses dihedral angle as an input feature, it does not distinguish between concave and convex edges.  This would explain why MeshCNN struggles to distinguish the subtractive extrusion classes (\textit{CutSide} and \textit{CutEnd}) from the more common additive extrusions (\textit{ExtrudeSide} and \textit{ExtrudeEnd}).  

The \textit{Fillet} and \textit{Chamfer} classes account for $10.22\%$ and $2.94\%$ of faces in the dataset.  Fillets are very distinctive features with smooth edges and typically cylindrical or toroidal  geometry.  Both BRepNet and the ECC have high IoU scores for this class ($97.57\%$ and $97.32\%$ respectively), demonstrating that these patterns in the input features are easy to spot using both architectures.  Faces in the \textit{Chamfer} class are much less distinctive.  They are often planar or conical and their edges can be either concave or convex.  Consequently BRepNet and the ECC achieve lower IoU scores of $81.80\%$  and $71.16\%$.   PointNet++ and MeshCNN also have higher IoU values for fillets than for chamfers.  For fillets these networks achieve IoUs of $49.31\%$ and $36.49\%$ respectively while for chamfers PointNet++ achieves only an IoU of $21.79\%$ and MeshCNN fails to detect any chamfer features.

Both \textit{Fillet} and \textit{Chamfer} faces tend to have relatively small areas and consequently are prone to under-sampling when fixed edge-count meshes and fixed size point clouds are generated, however the IoU values PointNet++ and MeshCNN achieve for \textit{Fillet} suggests this was not a major limiting factor for the relatively small solids in the \textit{Fusion 360 Gallery} segmentation dataset.  Comparing the results for \textit{Fillet} with the less distinctive \textit{CutSide} class we see that all network are more successful at detecting fillets, despite these faces having smaller areas and accounting for a similar fraction of the dataset.

The \textit{ExtrudeSide} and \textit{ExtrudeEnd} classes account for $50.60\%$ and $15.72\%$ of the dataset respectively.  All networks do well on these  classes.  It should be noted that for both BRepNet and the ECC the IoU achieved for the most common \textit{ExtrudeSide} class is smaller than for the more distinctive \textit{Fillet} class.